\documentclass{article}

    \usepackage[final]{neurips_2021}

\usepackage[utf8]{inputenc} 
\usepackage[T1]{fontenc}    
\usepackage{hyperref}       
\usepackage{url}            
\usepackage{booktabs}       
\usepackage{amsfonts}       
\usepackage{nicefrac}       
\usepackage{microtype}      
\usepackage{xcolor}         
\usepackage{amsmath}
\usepackage{graphicx}
\usepackage[ruled]{algorithm2e}
\usepackage{natbib}
\usepackage{bookmark}
\usepackage{xcolor}
\usepackage{amsmath}
\usepackage{svg}
\usepackage{wrapfig}
\usepackage{caption}
\usepackage{subcaption}
\usepackage{listings}
\usepackage{tikz}
\usepackage{tabto}
\usepackage{array}

\definecolor{bluecite}{HTML}{0875b7}

\hypersetup{
  colorlinks=true,
  citecolor=bluecite,
  linkcolor=magenta
}

\definecolor{codegreen}{rgb}{0,0.6,0}
\definecolor{codegray}{rgb}{0.5,0.5,0.5}
\definecolor{codepurple}{rgb}{0.58,0,0.82}
\definecolor{backcolour}{rgb}{0.95,0.95,0.92}

\SetAlgorithmName{Block}{List of protocols}

\lstdefinestyle{mystyle}{
    backgroundcolor=\color{backcolour},   
    commentstyle=\color{codegreen},
    keywordstyle=\color{magenta},
    numberstyle=\tiny\color{codegray},
    stringstyle=\color{codepurple},
    basicstyle=\ttfamily,
    breakatwhitespace=false,         
    breaklines=true,                 
    captionpos=b,                    
    keepspaces=false,                 
    numbers=left,                    
    numbersep=5pt,                  
    showspaces=false,                
    showstringspaces=false,
    showtabs=false,                  
    tabsize=2
}

\title{On pseudo-absence generation and machine learning for locust breeding ground prediction in Africa}

\author{
Ibrahim Salihu Yusuf\thanks{Correspondence: \texttt{i.yusuf@instadeep.com}} \\
InstaDeep
\And
Kale-ab Tessera \\
InstaDeep 
\And
Thomas Tumiel \\
InstaDeep 
\And
Zohra Slim \\
InstaDeep 
\And
Amine Kerkeni \\
InstaDeep 
\And
Sella Nevo \\
Google Research 
\And 
Arnu Pretorius \\
InstaDeep
}

\begin{document}
\maketitle

\begin{abstract}
    Desert locust outbreaks threaten the food security of a large part of Africa and have affected the livelihoods of millions of people over the years. 
    Furthermore, these outbreaks could potentially become more severe and frequent as a result of global climate change.
    Machine learning (ML) has been demonstrated as an effective approach to locust distribution modelling which could assist in early warning.
    However, ML requires a significant amount of labelled data to train. Most publicly available labelled data on locusts are presence-only data, where only the sightings of locusts being present at a particular location are recorded. Therefore, prior work using ML have resorted to pseudo-absence generation methods as a way to circumvent this issue and build balanced datasets for training. The most commonly used approach is to randomly sample points in a region of interest while ensuring these sampled pseudo-absence points are at least a specific distance away from true presence points. In this paper, we compare this random sampling approach to more advanced pseudo-absence generation methods, such as environmental profiling and optimal background extent limitation, specifically for predicting desert locust breeding grounds in Africa. Interestingly, we find that for the algorithms we tested, namely logistic regression, gradient boosting, random forests and MaxEnt, all popular in prior work, the linear logistic model performed significantly better than the more sophisticated ensemble methods, both in terms of prediction accuracy and F1 score. Although background extent limitation combined with random sampling seemed to boost performance for ensemble methods, no statistically significant differences were detected between the pseudo-absence generation methods used to train the logistic model. In light of this, we conclude that simpler approaches such as random sampling for pseudo-absence generation combined with linear classifiers such as logistic regression are sensible and effective for predicting locust breeding grounds across Africa.\footnote{Data preprocessing and modelling code: \href{https://github.com/instadeepai/locust-predict}{https://github.com/instadeepai/locust-predict}}
\end{abstract}

\section{Introduction}

Climatic conditions leading to cyclones and monsoons, causing heavy rains, prompted the recent 2019-2021 upsurge in desert locusts \citep{fao2021upsurge}.
These upsurges pose a significant threat to food security in affected areas, especially in the Northern parts of the African continent. 
Furthermore, the occurrence and severity of such upsurges could potentially be exacerbated by global climate change \citep{vallebona2008large, fao2016weather, zhang2019locust,salih2020climate}. 

The Food and Agriculture Organization (FAO) of the United Nations operate a sophisticated monitoring and early warning system for locust outbreaks \citep{fao2020early}.
The system relies on a range of technologies serving field survey operators, control centres and researchers \citep{cressman2008use}. 
In particular, remote sensing has become an invaluable component for early warning because of its usefulness in locust distribution modelling \citep{latchininsky2010locust, cressman2013role, latchininsky2013locusts, klein2021application}.
Female locusts typically lay their eggs in wet, warm soil and wingless nymph locusts, referred to as \textit{hoppers}, require specific vegetation nearby to sustain them before their wings develop \citep{symmons2001desert, fao2021standard}. 
This connection between certain environmental variables and locust behaviour makes it possible to attempt to model locust distribution through remote sensing combined with survey data \citep{piou2013coupling, escorihuela2018smos, piou2019soil, chen2020geographic, ellenburg2021detecting}.

Machine learning (ML) has been shown to be a valuable tool for species distribution modeling \citep{beery2021species}.
In particular, many recent papers have looked at using ML specifically for modelling locusts \citep{gomez2018machine,gomez2019desert,kimathi2020prediction,gomez2020modelling,gomez2021prediction}.
However, even when remote sensing is capable of providing useful features for such models, ML still heavily relies on large amounts of labelled data for training.
Currently, the FAO provides many years worth of labelled data on locusts hosted through their Locust Hub.\footnote{\href{https://locust-hub-hqfao.hub.arcgis.com/}{https://locust-hub-hqfao.hub.arcgis.com/}}
This is an extremely useful resource and contains recorded sightings of locusts in various phases and stages of their lifecycle. 
That said, survey teams in general only record the presence of locusts and rarely their absence. 
This is typical of many ecological surveys and data of this kind are referred to as \textit{presence-only} data.
To overcome the lack of negative labels when training ML models, past work have made use of \textit{pseudo-absence} generation \citep{gomez2018machine,gomez2020modelling,gomez2021prediction}.
A commonly used approach is to randomly sample points in a region of interest while ensuring that pseudo-absence points are sampled a minimum distance away from any true presence points \citep{barbet2012selecting}. 
More advanced pseudo-absence generation methods also exist, such as environmental profiling and background extent limitation techniques \citep{iturbide2015framework}.

In this paper, we compare different pseudo-absence generation methods used in conjunction with ML, including methods such as random sampling, environmental profiling and background extent limitation, specifically when modelling the desert locust species in Africa. 
We focus on ML algorithms commonly used in prior work: logistic regression (LR), gradient boosting (XGBoost), random forests (RF) and maximum entropy (MaxEnt -- a presence-background modelling approach\footnote{Presence-background approaches make use of only labelled presence data and points sampled across the entire study area of interest, referred to as \textit{background data}, without using any absence or pseudo-absence data. Even though MaxEnt does not rely on pseudo-absence generation, we include it in this work as a strong baseline for comparison.}).
We train each algorithm on specific environmental variables combined with presence labels from the FAO's Locust Hub. 
For LR, XGBoost and RF we generate pseudo-absence labels using each of the above-mentioned pseudo-absence generation methods.
Our results show LR performing significantly better in terms of prediction accuracy and F1 score compared to the ensemble methods XGBoost and RF as well as MaxEnt.
For LR training, there are no statistically significant differences between the pseudo-absence generation methods, whereas for the ensemble methods, random sampling combined with background extent limitation significantly improved performance.
We therefore conclude, as well as simultaneously validate, the approach of random sampling for pseudo-absence generation, used and studied by \cite{gomez2018machine,gomez2019desert} and \cite{barbet2012selecting} to be effective. 
However, in contrast to these works, we find the simpler linear model LR, as opposed to the RF, to be more suitable when used for breeding ground prediction over a large region of Africa.

\section{Methodology}

We are interested in testing the effectiveness of different pseudo-absence generation methods combined with ML for modelling locusts. 
Here we discuss our methodology concerning our data, including our choice of environmental variables and preprocessing.
We explain the different pseudo-absence generation methods and ML algorithms in more detail and formally state our research hypothesis.

\textbf{Data.} We focus on modelling desert locusts over the entire affected region of the African continent (we provide the full list of countries in the supplementary material (SM)). We use the FAO's Locust Hub observation data in this study, as it contains the geo-locations of areas where locusts were observed, type of locust observed and some environmental conditions. It has a temporal range of 1985 to 2021. 
The observations were enriched with environmental data from NASA\footnote{\href{https://disc.gsfc.nasa.gov/datasets/GLDAS\_NOAH025\_3H\_2.1/summary}{https://disc.gsfc.nasa.gov/datasets/GLDAS\_NOAH025\_3H\_2.1/summary}} and ISRIC SoilGrids\footnote{\href{https://data.isric.org/geonetwork/srv/eng/catalog.search\#/home}{https://data.isric.org/geonetwork/srv/eng/catalog.search\#/home}} respectively (names and descriptions of each variable are provided in SM). 
Our variables include soil characteristics such as moisture, profile (type) and temperature as well as air pressure, humidity and surface level temperature.
The environmental data from NASA have a temporal range of 2000 to 2021, while the soil profile information is non-temporal. The three datasets were combined by selecting a region of temporal overlap from 2000 to 2021. 
As in \cite{gomez2018machine}, we use hopper presence/absence as a proxy for locust breeding grounds.
Given that the maximum time period between the start of egg laying to the end of the hopper phase is approximately 95 days and that hoppers are not able to fly, they remain close to the breeding ground and therefore act as a good proxy.
Therefore, we use a 95-day history for all temporal variables and predict hopper presence 7-days into the future as a proxy for predicting the presence of potential future breeding grounds.
The resulting dataset was split into train and test sets based on time; the period from 2000-2014 was used for training, while the period from 2015-2021 was used for testing. 
Pseudo-absence generation was also performed separately on each split ensuring a balanced distribution of hopper presence/absence. 
After generating pseudo-absences the train set was further split into a smaller train set and a validation set for parameter tuning in the ratio of 80:20.
To ensure that different algorithm and pseudo-absence pairs could be tested fairly and on the same test set, we constructed pseudo-absences for the test set by randomly sampling a balanced mixture of test pseudo-absence points generated by the different methods operating on the test set presence points. For more details regarding the dataset, we refer the reader to the SM.

\begin{figure}
    \centering
    \includegraphics[width=0.99\textwidth]{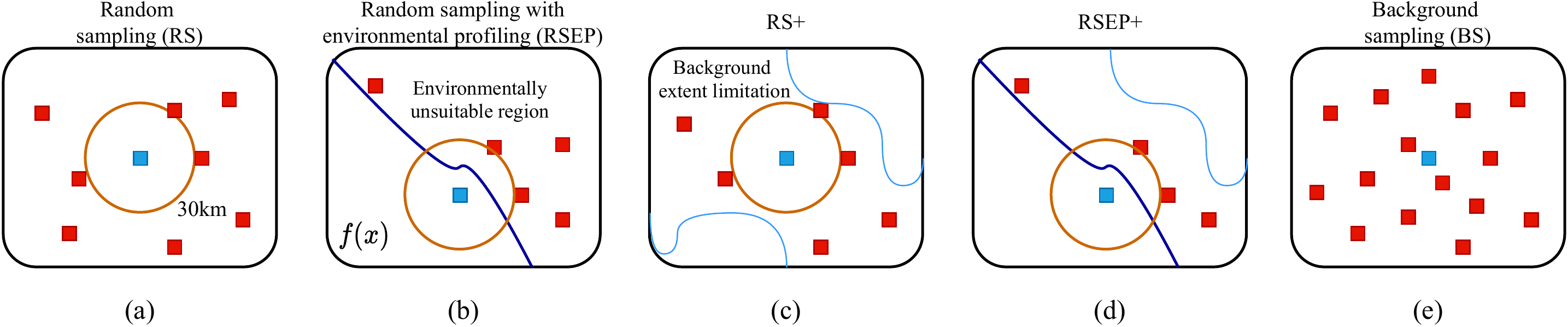}
    \caption{ \textit{Illustrations of pseudo-absence and background data generation}. Blue square is a presence point and red squares represent generated pseudo-absence points. \textbf{(a)} Random sampling (RS). Orange circle is the exclusion buffer. \textbf{(b)} Random sampling with environment profiling (RSEP). Dark blue line shows the boundary between environmentally suitable and unsuitable regions. \textbf{(c)} Random sampling with background extent limitation (RS+). Light blue lines show excluded background regions. \textbf{(d)} Random sampling with environment profiling and background extent limitation (RSEP+). \textbf{(e)} Background sampling (BS) for presence-background modelling. It is important to note that for BS the red points do not necessarily correspond to absence but could represent both presence and absence.} 
    \label{fig: pa gen}
    \end{figure}

\textbf{Pseudo-absence generation.} We consider four different approaches for pseudo-absence generation detailed in \cite{iturbide2015framework} and depicted in Figure \ref{fig: pa gen} (a)-(d): 
\begin{enumerate}
    \item \textit{Random sampling (RS)}: Pseudo-absences are sampled at random across all points in the study area that are not within some minimum selected distance to any presence point. The minimum distance between a presence and any absence point is referred to as the \textit{exclusion buffer} and is set to 30km for all methods. 
    \item \textit{Random sampling with environmental profiling (RSEP)}: The RSEP method is aimed at defining the environmental range of the background from which pseudo-absences are sampled. Environmentally unsuitable areas for locusts are defined using a presence-only profiling algorithm (one-class SVM) trained on soil moisture. Once these unsuitable regions have been established, pseudo-absence points are randomly sampled from within them.
    \item \textit{RS with background extent limitation (RS+)}: Pseudo-absences are sampled at random within a limited background extent not including the full study region. This optimum limited background is determined by a multi-step process as outlined in \citep{iturbide2015framework}.\footnote{We perform all pseudo-absence generation using the \texttt{mopa} R package \citep{iturbide2015framework}.}
    \item \textit{RSEP with background extent limitation (RSEP+)}: This method is similar to the above RS+, but instead of using unconditioned random sampling within the limited background extent, samples are only within environmentally unsuitable regions identified through profiling.
\end{enumerate}

As a strong baseline we compare these pseudo-absence generation methods to only using presence-background data (see Figure \ref{fig: pa gen} (e)) modelling where background samples are generated over the entire study region of interest without any constraint on their location with respect to presence points.

\textbf{ML algorithms.} We consider the following algorithms for comparison: logistic regression (LR), gradient boosting (XGBoost) \citep{freund1997decision,friedman2001greedy}, random forests (RF) \citep{breiman2001random} and maximum entropy (MaxEnt) \citep{phillips2006maximum}. We provide hyperparameter details in SM. 

\textbf{Hypothesis.} Our null hypothesis, $H_0$, is that of no difference between the mean performances of the different pseudo-absence generation methods across all the algorithms tested including the mean performance of MaxEnt using only presence-background data. 
Specifically, let $G = \{rs, rsep, rs+, rsep+\}$, $A = \{lr, xgboost, rf\}$ and let $\mu^g_a$ represent the mean performance for pseudo-absence generation method $g \in G$, used when training algorithm $a \in A$. Then the null hypothesis is given by
\begin{align}
    H_0: \mu^g_a = \mu^{g^\prime}_{a^\prime} = \mu^{BS}_{MaxEnt} \hspace{1cm} \forall g, g^\prime \in G, a,a^\prime \in A | g \neq g^\prime \land a \neq a^\prime
\end{align}
We are required to reject $H_0$ to have evidence that there are indeed differences between the pseudo-absence generation methods used for training the different algorithms as well as the presence-background MaxEnt approach. Our significance level, i.e.\ the point at which the probability of equal mean performances under the null hypothesis is low enough to reject the null hypothesis, is set to $\alpha = 0.05$. If $H_0$ is rejected, we can continue with more specific pairwise tests, with appropriate $p$-value adjustments to control for the \textit{family-wise error} \citep{demvsar2006statistical}.

\section{Experiments}

\begin{figure}
    \centering
    \includegraphics[width=0.19\textwidth]{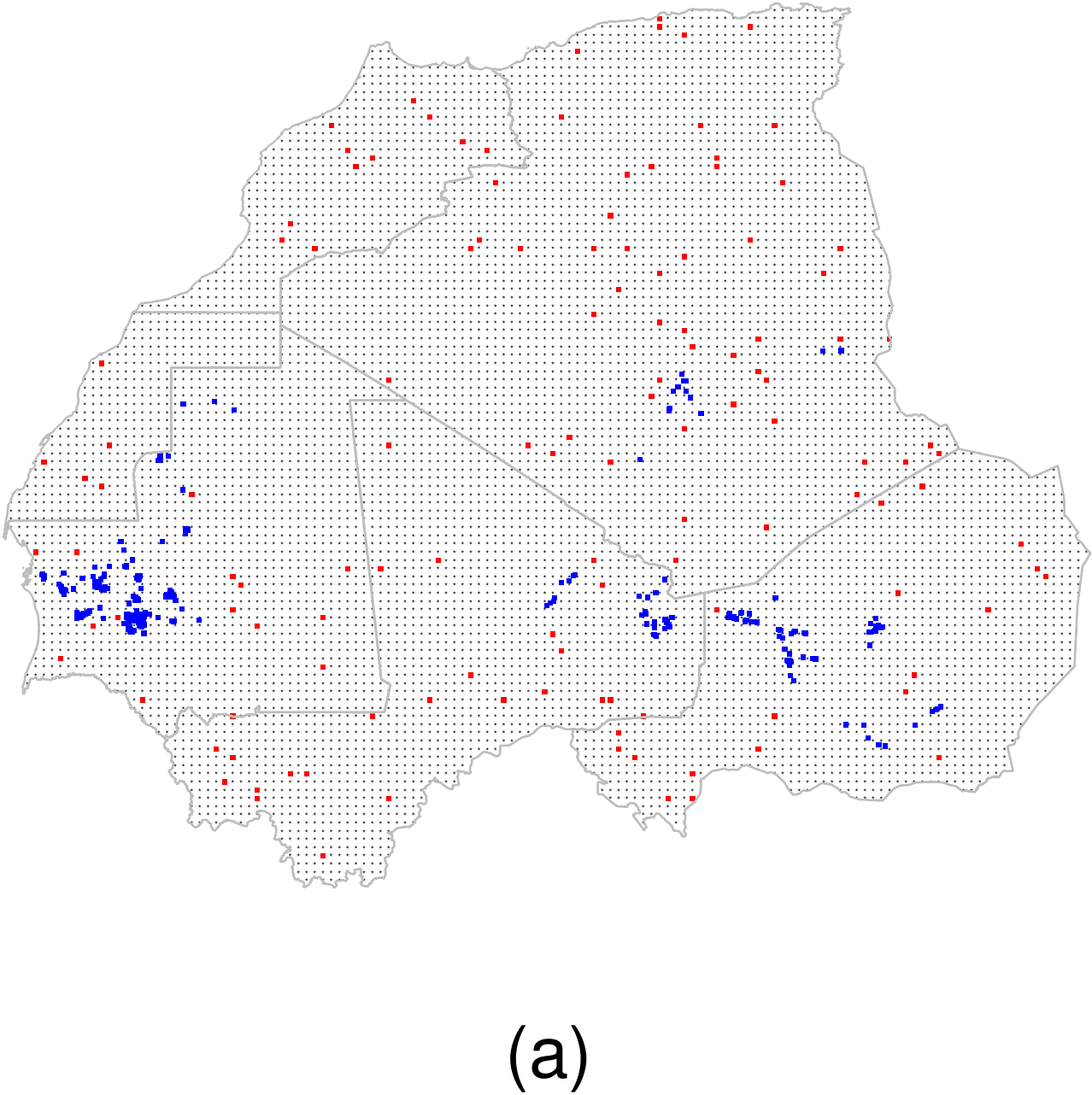}
    \includegraphics[width=0.19\textwidth]{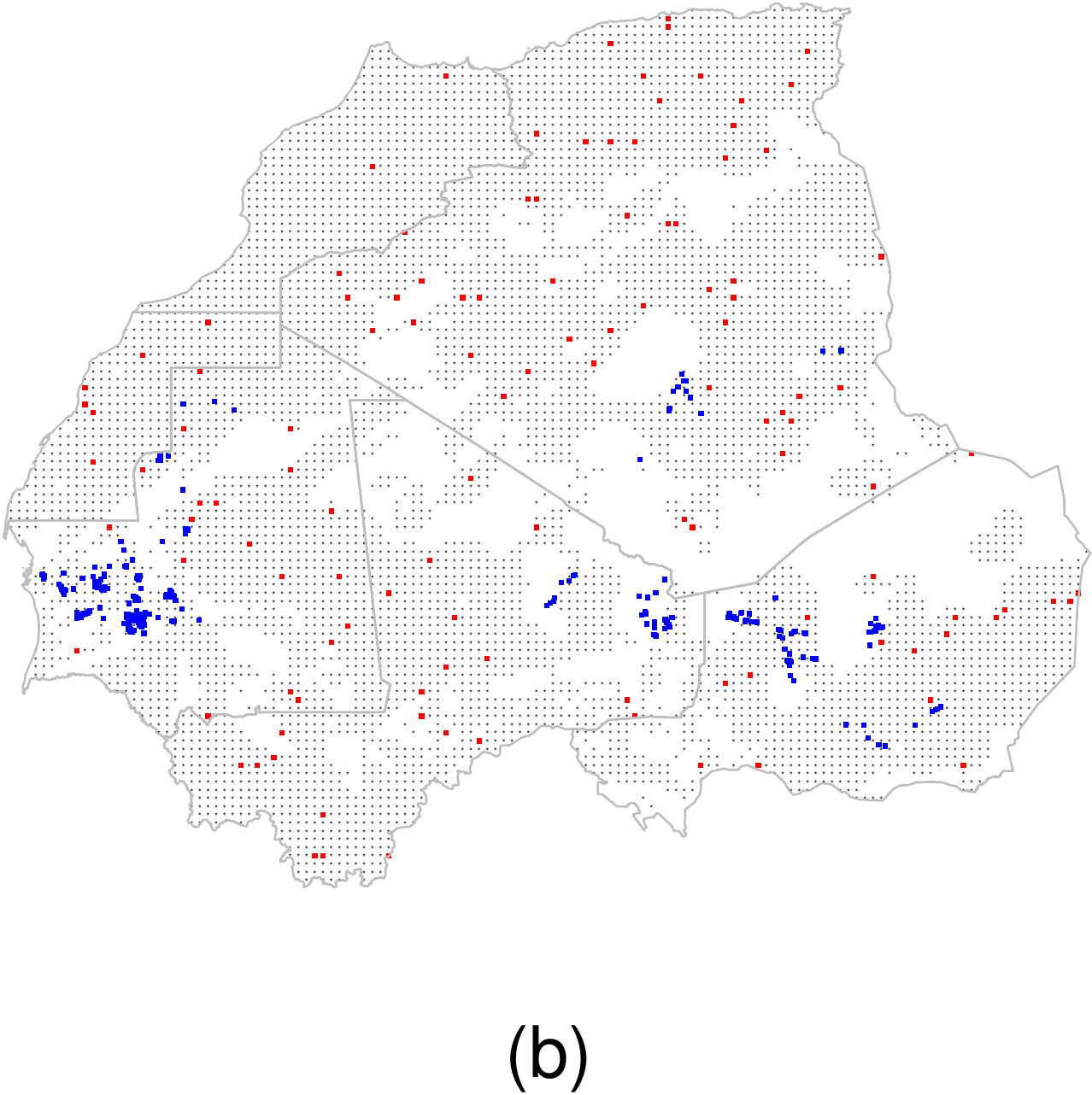}
    \includegraphics[width=0.19\textwidth]{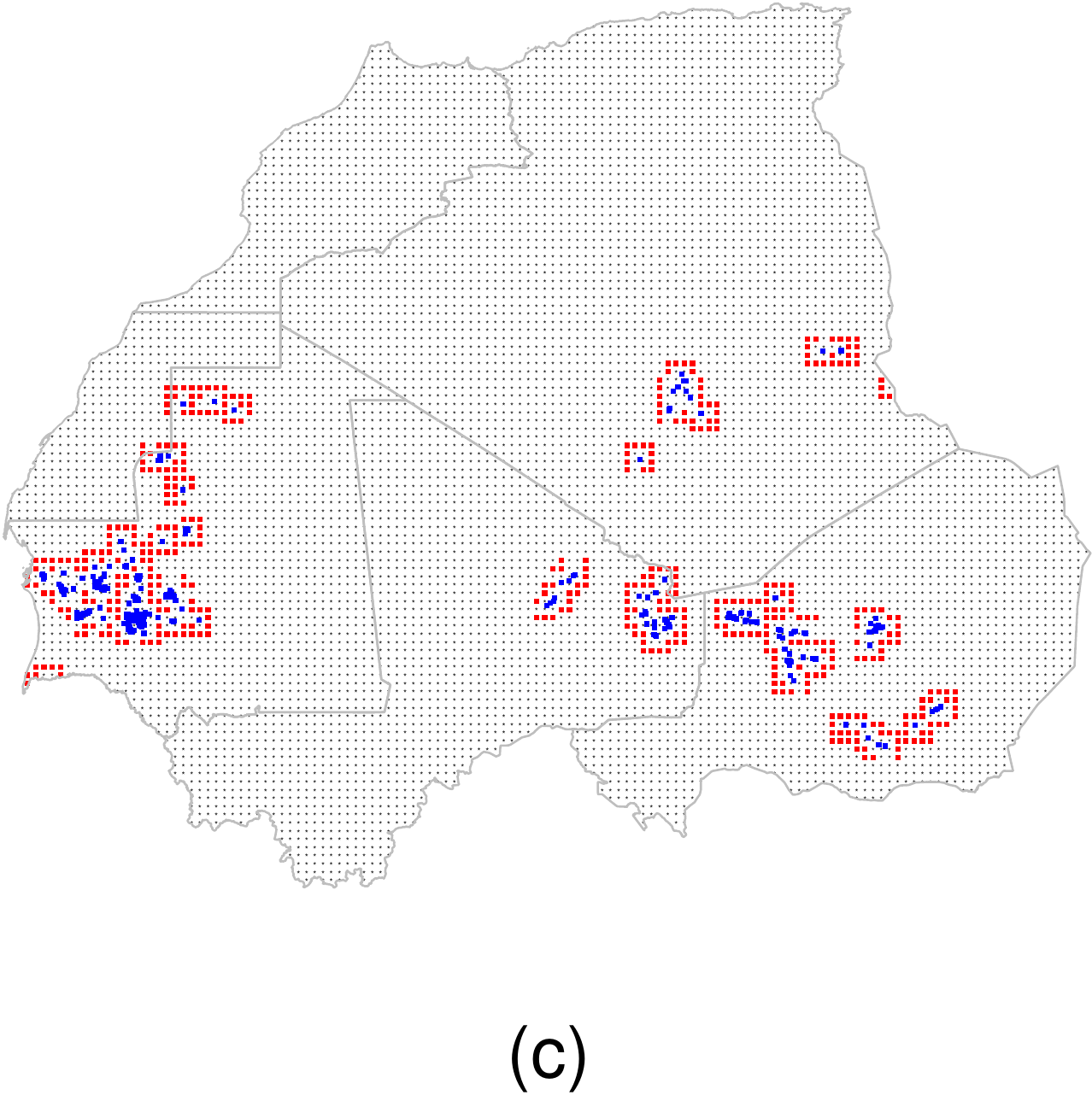}
    \includegraphics[width=0.19\textwidth]{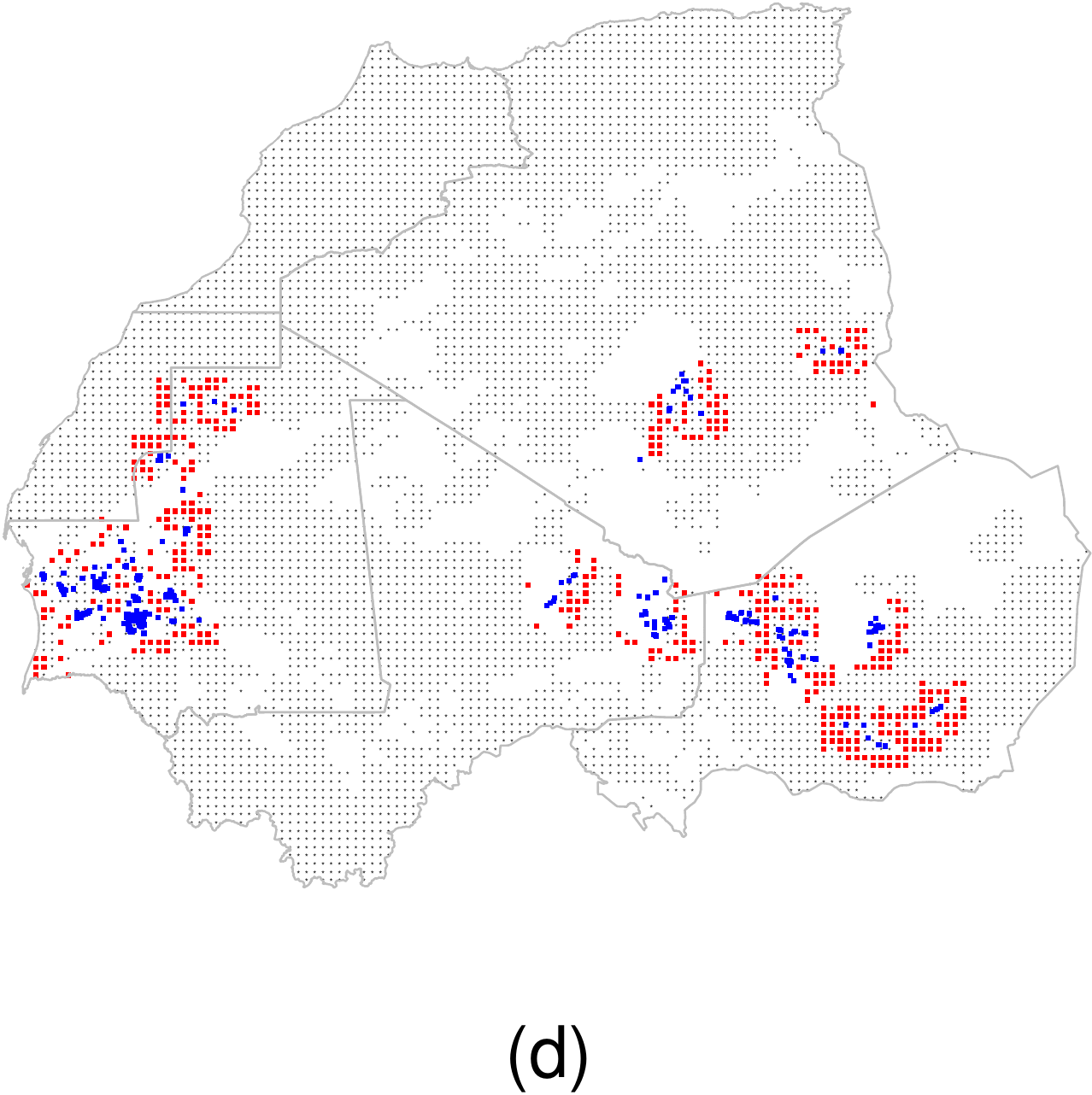}
    \includegraphics[width=0.19\textwidth]{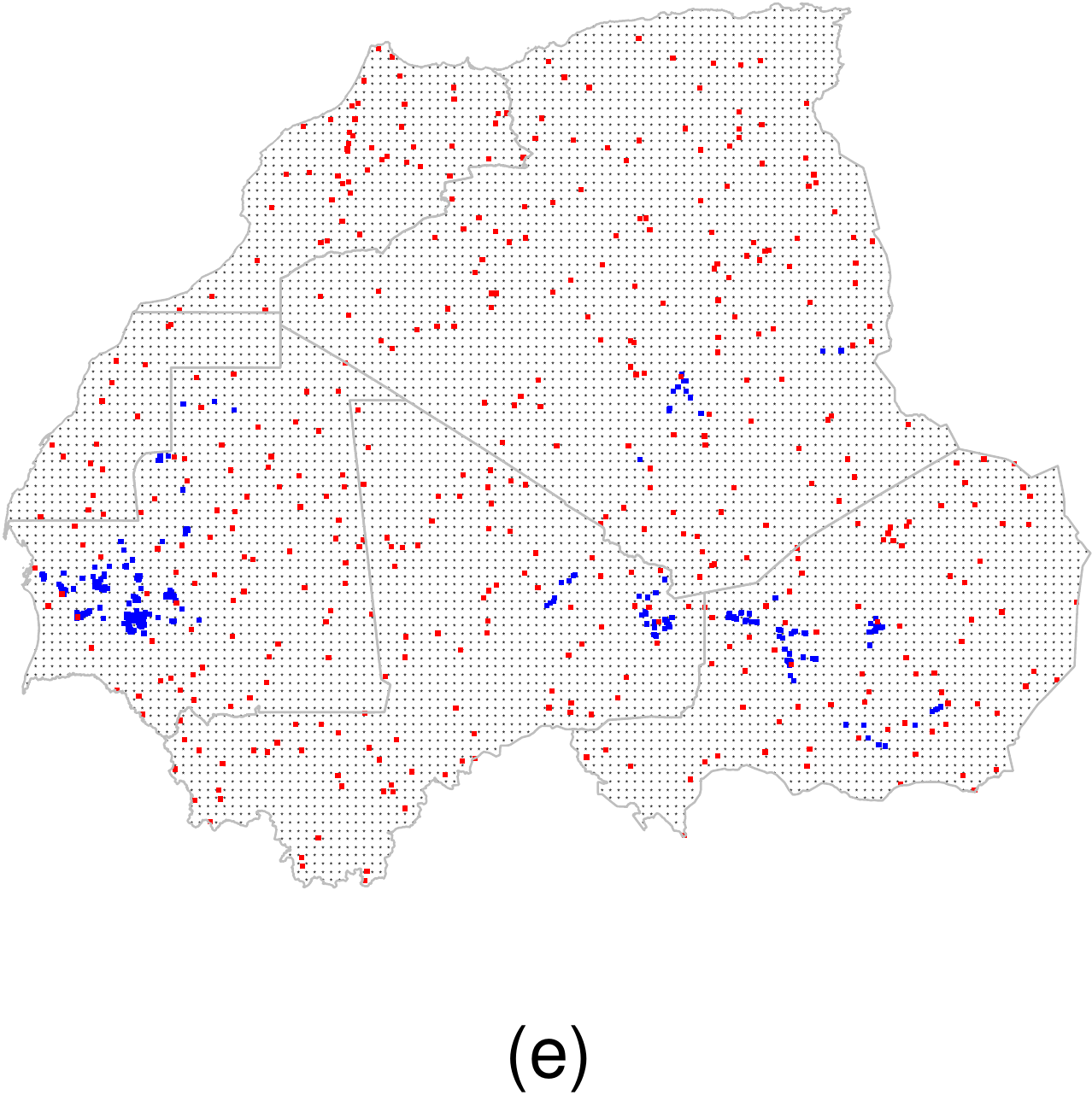}
    \caption{ \textit{Pseudo-absence and background data generation: an example on a subset of African countries Niger, Mauritania, Mali, Algeria, Western Sahara and Morocco for November 2003}. \textbf{(a)} Random sampling (RS). \textbf{(b)} Random sampling with environment profiling (RSEP). White regions indicate environmentally suitable regions as identified through environmental profiling, i.e.\ these are the regions where pseudo-absence points should not be sampled. \textbf{(c)} Random sampling with background extent limitation (RS+). \textbf{(d)} Random sampling with environment profiling and background extent limitation (RSEP+). \textbf{(e)} Background sampling (BS). Note that for BS the red points do not necessarily correspond to absence but could represent both presence and absence.} 
    \label{fig: pa gen for North Africa}
\end{figure}

As an example of the different generation methods, in Figure \ref{fig: pa gen for North Africa}, we show the pseudo-absence points generated by each method in the countries Niger, Mauritania, Mali, Algeria, Western Sahara and Morocco for November 2003. 

\textbf{Results.} The mean accuracy and F1 score over 100 runs for each algorithm and generation method is shown in Table \ref{tab: acc}. In general, the performances are respectable. However, interestingly, the linear model LR is seen to outperform the more sophisticated ensemble methods, XGBoost and RF, across all generation methods, as well as MaxEnt, both in terms of absolute mean accuracy and F1 score. Next, we ascertain the statistical significance of these results.

\begin{table}
    \caption{Performance comparison between generation methods and ML algorithms. Bold values indicate top performance across generation methods for a specific algorithm.}
    \label{tab: acc}
    \centering
    \resizebox{0.8\columnwidth}{!}{%
    \begin{tabular}{llccc|c} \toprule
          & & LR & XGBoost & RF &  MaxEnt \\ \midrule
           & RS & $0.8499\pm0.0021$ & $0.7320\pm0.0016$ & $0.6623\pm0.0112$ & \\
          Accuracy & RSEP & $\mathbf{0.8541\pm0.0020}$ & $0.7020\pm0.0015$ & $0.6706\pm0.0133$ & $0.7418\pm0.0019$ \\ 
           & RS+ & $0.8417\pm0.0020$ & $\mathbf{0.7580\pm0.0021}$ & $\mathbf{0.7855\pm0.0176}$ &  \\
           & RSEP+ & $0.8530\pm0.0022$ & $0.7423\pm0.0021$ & $0.7712\pm0.0138$  & \\ \midrule
           & RS & $0.9064\pm0.0012$ & $0.8157\pm0.0009$ & $0.7495\pm0.0104$ &  \\
         F1  & RSEP & $\mathbf{0.9098\pm0.0011}$ & $0.7895\pm0.0008$ & $0.7576\pm0.0121$ & $0.5437\pm0.0035$ \\
           & RS+ & $0.9008\pm0.0011$ & $\mathbf{0.8448\pm0.0012}$ & $\mathbf{0.8660\pm0.0121}$  & \\
           & RSEP+ & $0.9093\pm0.0013$ & $0.8293\pm0.0012$ & $0.8527\pm0.0101$ &  \\ \bottomrule
    \end{tabular}%
    }
\end{table}


\textbf{Statistical analysis.} We test $H_0$ using the Friedman aligned ranks test \citep{friedman1937use} and find that we can easily reject the null hypothesis of no difference with $p$-value $< 2 \times 10^{-16}$. 
Having rejected $H_0$, we perform additional pairwise tests using the $p$-value adjustment procedure for multiple testing from \cite{holm1979simple}.
For XGBoost and RF significant differences are detected between the generation methods (all $p$-values $< 2 \times 10^{-16}$)\footnote{We conducted our statistical testing in R using the \texttt{stats} library where $2 \times 10^{-16}$ is considered the minimum $p$-value. Values smaller than this are indicated with the `$<$' symbol in printed output.}, providing evidence that background extent limitation can be of benefit to these ensemble methods. 
For LR, the differences between the generation methods are not significant ($p$-values provided in SM). 
Given the seemingly absolute superiority in performance from LR with no difference between generation methods, we conduct a pairwise test between only the top performing algorithms and generation pairs and LR using the simplest random sampling. 
More specifically, we test the null hypothesis $H^a_0: \mu^{rs}_{lr} = \mu^{rs+}_{rf} = \mu^{rs+}_{xgboost} = \mu_{maxent}$. 
We find that the difference between LR and the other algorithms to be significant ($p < 2 \times 10^{-16}$). 

\begin{figure}
  \centering
  \includegraphics[width=0.35\textwidth]{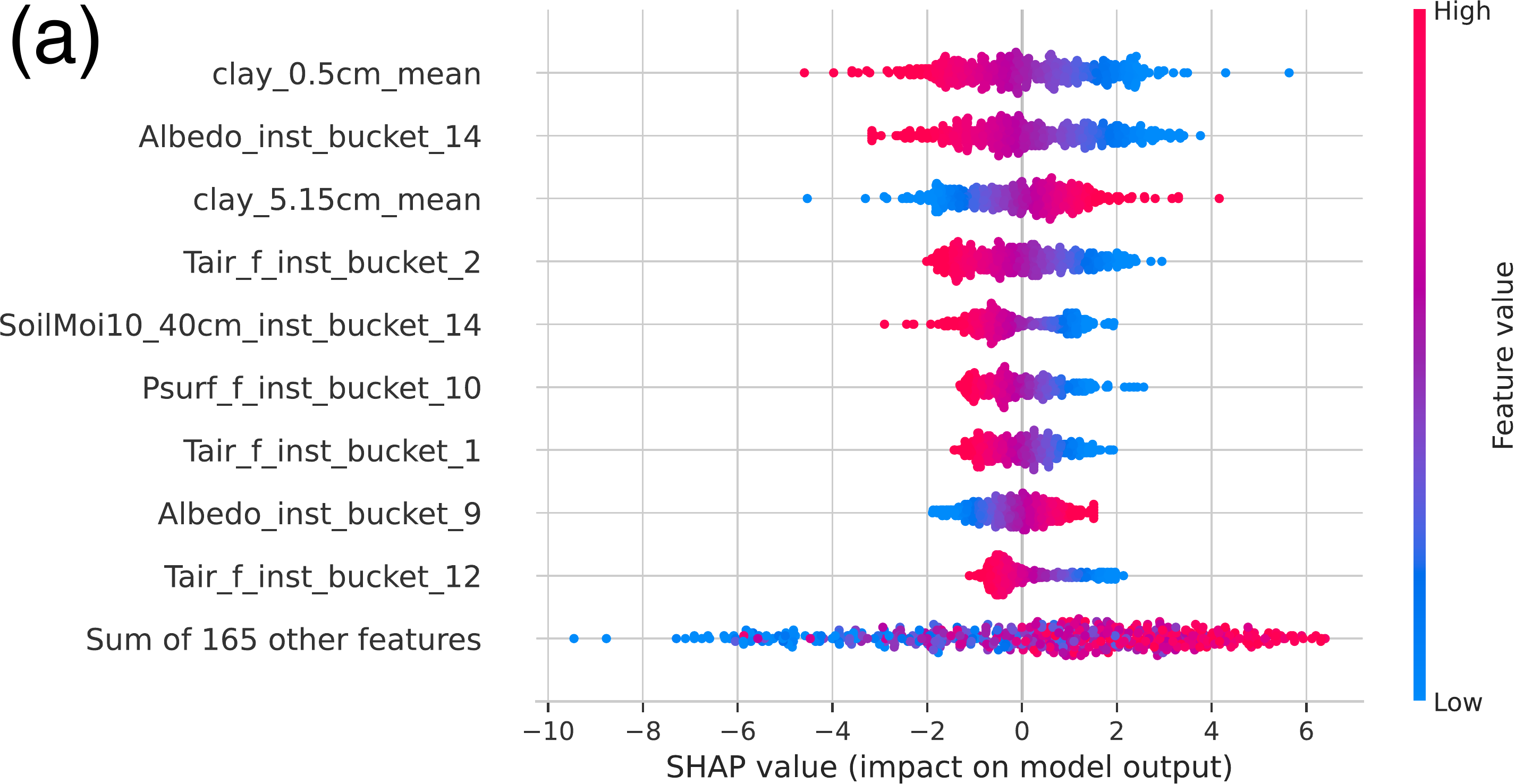} \hspace{0.5cm}
  \includegraphics[width=0.35\textwidth]{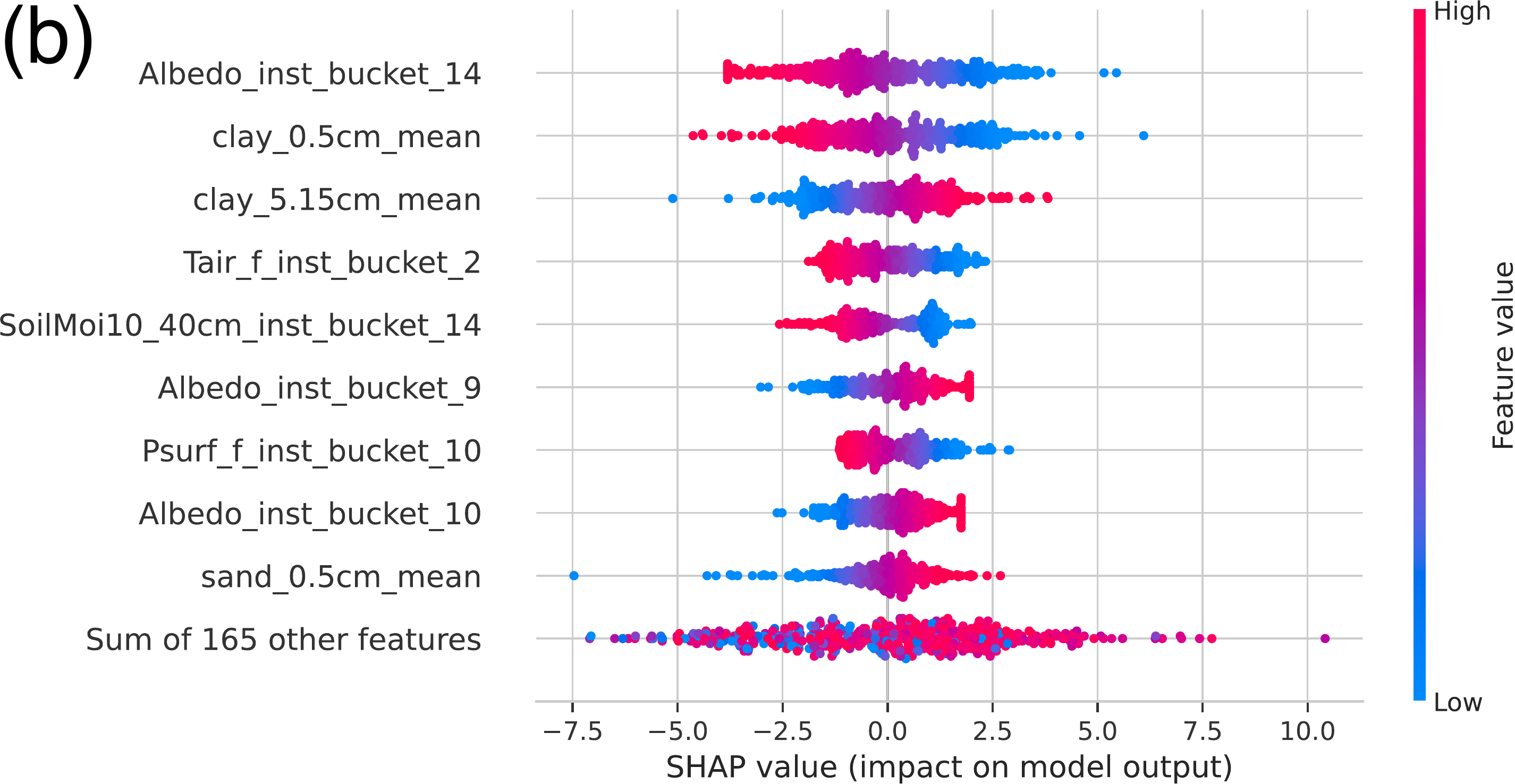}
  \caption{ \textit{SHAP analysis for logistic regression.} An interpretation of the logistic regression model on the different pseudo-absence generation methods using Shapley additive explanations \citep{NIPS2017_7062}. \textbf{(a)} Random sampling (RS). \textbf{(b)} Random sampling with environment profiling (RSEP).} 
  \label{fig: interpret}
\end{figure}

\textbf{Interpretation.} In Figure \ref{fig: interpret}, we provide a SHAP analysis \citep{NIPS2017_7062} for variable importance of the LR model between using RS or RSEP. We find that only a few of the total 174 features are important for prediction. For both methods features such as \texttt{Albedo\_inst\_bucket\_14} and \texttt{clay\_0.5cm\_mean} roughly amount to the same predictive effect as the bottom 165 variables combined. This is in contrast to the ensemble methods where feature importance is far more spread out. Therefore, we suspect the LR is performant due to its ability to avoid overfitting to noisy features.

\section{Discussion}

Our study provides some evidence for the suitability of using random sampling for pseudo-absence and logistic regression for predicting desert locust breeding grounds in Africa. 
We note that a detailed analysis of pseudo-absence generation appeared in \cite{barbet2012selecting}, but was performed on synthetic data, whereas we are primarily interested in locust modeling and considered more modern methods such as background extent limitation. Finally, although we find linear models to perform well in our study, recent papers have started to use deep learning for locust modeling \citep{samil2020predicting, tabar2021plan}, moving towards more sophisticated ML model architectures and leveraging promising new sources of data such as those generated by the eLocust3m app \citep{plant2020elocust} to good effect.
We hope to compare with and potentially contribute to these approaches in future work.

\newpage

\begin{ack}
This research was supported in part through computational resources provided by Google.
\end{ack}

\bibliographystyle{IEEEtranN}
\bibliography{references}

\newpage

\section*{Supplementary Material}

\textbf{Countries in study region.} Mauritania, Mali, Egypt, Morocco, Algeria, Sudan, South Sudan, Niger, Eritrea, Senegal, Libya, Western Sahara, Uganda, Tunisia, Cape Verde, Chad, Ethiopia, Kenya, Somalia, and Djibouti.

\textbf{Dataset details -- environmental variables, splits, sizes and features.} Temporal variables were extracted from NASA Global Land Data Assimilation System Version 2.1 (GLDAS-2.1) Noah Land Surface Model with a temporal resolution of 3 hours and spatial resolution of 0.25 x 0.25 degrees. \href{https://disc.gsfc.nasa.gov/datasets/GLDAS\_NOAH025\_3H\_2.1/summary}{https://disc.gsfc.nasa.gov/datasets/GLDAS\_NOAH025\_3H\_2.1/summary} 

\begin{table}[h]
    \caption{Environmental variables and their descriptions}
    \label{tab: env vars 1}
    \centering
    \begin{tabular}{ll} \toprule
        Name & Description \\ \midrule
        AvgSurfT\_inst & Instantaneous average surface skin temperature (K) \\ 
        Albedo\_inst & Instantaneous albedo (\%) \\ 
        SoilMoi0\_10cm\_inst & Instantaneous soil moisture 0-10cm (kg m-2) \\ 
        SoilMoi10\_40cm\_inst & Instantaneous soil moisture 10-40cm (kg m-2)\\ 
        SoilTMP0\_10cm\_inst & Instantaneous soil temperature 0-10cm (K)\\ 
        SoilTMP10\_40cm\_inst & Instantaneous soil temperature 0-10cm (kg m-2)\\ 
        Tveg\_tavg & 3-hour averaged Transpiration (W m-2)\\ 
        Wind\_f\_inst & Instantaneous wind speed (m s-1)\\ 
        Rainf\_f\_tavg & 3-hour averaged total precipitation rate (kg m-2 s-1)\\ 
        Tair\_f\_inst & Instantaneous air temperature (K)\\ 
        Qair\_f\_inst & Instantaneous specific humidity (kg kg-1)\\ 
        Psurf\_f\_inst & Instantaneous surface pressure (Pa)  \\ \bottomrule
    \end{tabular}
\end{table}

Soil profile variables were extracted from International Soil Reference and Information Centre (ISRIC) SoilGrids250m 2.0 data product. \href{https://data.isric.org/geonetwork/srv/eng/catalog.search\#/home}{https://data.isric.org/geonetwork/srv/eng/catalog.search\#/home}

\begin{table}[h]
    \caption{Environmental variables and their descriptions}
    \label{tab: env vars 2}
    \centering
    \begin{tabular}{ll} \toprule
        Name & Description \\ \midrule
        sand\_0.5cm\_mean & Average sand content between 0-5cm (g/kg) \\
        sand\_5.15cm\_mean & Average sand content between 5-15cm (g/kg) \\
        clay\_0.5cm\_mean & Average clay content between 0-5cm (g/kg) \\ 
        clay\_5.15cm\_mean & Average clay content between 5-15cm (g/kg) \\
        silt\_0.5cm\_mean & Average silt content between 0-5cm (g/kg) \\
        silt\_5.15cm\_mean & Average silt content between 5-15cm (g/kg) \\ \bottomrule
    \end{tabular}
\end{table}

\begin{table}[h]
    \caption{Dataset splits, sizes and number of features}
    \label{tab: env vars details}
    \centering
    \begin{tabular}{llllll} \toprule
        Split & Feature type(s) & Number of features & Presence & Pseudo-Absence & Total \\ \midrule
        Train &  Numeric & 174 & 17007 & 9251 & 26258 \\
        Val & Numeric & 174 & 4206 & 2359 & 6565 \\
        Test & Numeric & 174 & 5842 & 1238 & 7080 \\ \bottomrule
    \end{tabular}
\end{table}

We used a total of 12 temporal and 6 non-temporal variables. For each temporal variable, we retrieved a 95-day history and removed the last 7 days (including the observation day), to ensure that we predict 7 days ahead. Futhermore, we took each 89-day window and bucketised over time by computing the mean value for every 6-day window. We dropped the last window, which had less than 6 days. After doing this we arrived at 14 windows for each variable, so that in total we had 168 (14*12) temporal features. In addition to 6 non-temporal features the total number of features was 174. 

\textbf{ML model hyperparameters.} We performed manual hyperparameter tuning on the validation set resulting in the follow values: a maximum tree depth of 4 for XGBoost and 15 for RF. The number of random variables used for node splitting in RF was $\sqrt{p}$, where $p$ is the total number of variables. For our MaxEnt model, we used a linear feature class and a regularization factor of 1. These hyperparameters were found using a grid search on the training data over the available feature classes (linear, quadratic, product, threshold and hinge) and regularization factor in the range [0.1,1], with a step size of 0.1. We used the MaxNet library from \cite{phillips2017opening} for modeling.

\end{document}